\documentclass[journal,twoside,web]{IEEEtran}
\usepackage{silence}
\usepackage[font=scriptsize]{caption}
\usepackage{booktabs, makecell, multirow, tabularx}
\usepackage{cite}
\usepackage{amsmath,amssymb,amsfonts}
\usepackage{algorithmic}
\usepackage{graphicx}
\usepackage{textcomp}
\usepackage{hyperref}
\usepackage[utf8]{inputenc}
\usepackage[T1]{fontenc}
\usepackage{color,soul}
\usepackage{times}
\usepackage[nointegrals]{wasysym} 
\usepackage{epsfig}
\usepackage{multicol,multirow}
\usepackage{booktabs}
\usepackage{bigstrut}
\usepackage{floatrow}
\usepackage{rotating}
\usepackage{adjustbox}
\usepackage{mathtools}
\usepackage{longtable}
\usepackage{array}
\usepackage{pifont}

\def\BibTeX{{\rm B\kern-.05em{\sc i\kern-.025em b}\kern-.08em
    T\kern-.1667em\lower.7ex\hbox{E}\kern-.125emX}}
\begin{document}
\title{Progressive Class-Wise Attention (PCA) Approach for Diagnosing Skin Lesions}
\author{Asim~Naveed, Syed S. Naqvi, Tariq M. Khan, Imran Razzak
\thanks{Asim Naveed and Syed S. Naqvi are with the Department of Electrical and Computer Engineering, COMSATS University Islamabad (CUI), Islamabad 45550, Pakistan}
\thanks{Tariq M. Khan and Imran Razzak are with the School of Computer Science and Engineering, UNSW, Sydney, Australia}}
\maketitle
\begin{abstract}
Skin cancer holds the highest incidence rate among all cancers globally. The importance of early detection cannot be overstated, as late-stage cases can be lethal. Classifying skin lesions, however, presents several challenges due to the many variations they can exhibit, such as differences in colour, shape, and size, significant variation within the same class, and notable similarities between different classes. This paper introduces a novel class-wise attention technique that equally regards each class while unearthing more specific details about skin lesions. This attention mechanism is progressively used to amalgamate discriminative feature details from multiple scales. The introduced technique demonstrated impressive performance, surpassing more than 15 cutting-edge methods including the winners of HAM1000 and ISIC 2019 leaderboards. It achieved an impressive accuracy rate of 97.40
\end{abstract}


\section{Introduction}

The skin makes up a significant amount of a person's body, accounting for almost 16\% of total body weight. The three layers of skin are the epidermis, dermis, and subcutaneous tissues, also referred to as the hypodermis. Thermoregulation, metabolism, sensory perception, and protection are among the most significant skin functions. The skin protects the body against external invasions by acting as a barrier. Blood vessels, lymphatic vessels, nerves, and muscles are all found in the skin, enabling sweat to pass through the pores, regulating body temperature, and protecting the body from external elements. Excessive sun exposure, heat and cold, chemicals, medications and other factors can cause skin diseases \cite{razzak2020skin}.

Dermatology is challenging due to the complexity of the diagnostic processes required for skin diagnosis. Due to the complexity of the problem, which involves significant variations in skin tone and other skin diseases, it becomes even more challenging to diagnose the precise forms of the disease. There are currently more than 200 known skin diseases, and the number grows day by day. Nearly 80\% of the population has some type of skin disease \cite{9264664}.

In the past three decades, skin cancer has become one of the most lethal and fast-spreading malignancies in the world \cite{siegel2021cancer}.
Skin cells are classified into three types: squamous cells, basal cells, and melanocyte cells. Melanocytic and non-melanocytic skin cancers are two different forms. Doctors frequently misclassify malignant melanoma as benign since it is challenging to distinguish melanoma tumours from benign ones. To reduce the risk of death, it is crucial to recognise the right form of cancer at an early stage \cite{saeed2021skin}. Melanoma is the most prevalent cancer that develops in the melanocyte cells and rapidly spreads throughout the body \cite{okur2018survey}. It is more fatal than basal and squamous cell carcinoma. Any part of the human body can be affected by it. It is most typically found in sun-exposed regions like hands, face, neck, lips, and other similar locations. Melanoma skin cancer is treatable if it is found in its early stages; if not, it can spread to other body areas and result in a painful death \cite{khan2019classification}. Melanoma skin cancer makes up just 5\% of the occurrences of skin cancer overall, but it is more deadly, according to the American Cancer Society \cite{dildar2021skin}. 
Doctors or dermatologists frequently utilize the ABCDE method (A for asymmetry, B for border, C for colour, D for diameter, and E for evolution) for skin lesions diagnosis. Dermatologists employ the ABCDE assessment criteria to identify skin cancer in its early stages \cite{healsmith1994evaluation}. Owing to the quantitative nature of the problem, there is a lot of interest in the creation of semi or completely autonomous computer-aided diagnostic (CAD) systems. These systems may be applied in different settings such as decision support in screening or as an alternative opinion \cite{mahbod2020transfer}.

Medical image segmentation is a critical tool in healthcare, greatly improving diagnostic and therapeutic processes \cite{khawaja2019improved, khawaja2019multi, khan2021residual, khan2022width, khan2023neural}. Offering a transparent view of internal bodily formations, such as tumours or lesions, it equips doctors with the capability to make accurate and prompt diagnoses \cite{iqbal2022recent, khan2023simple, naqvi2023glan, khan2023retinal}. Beyond visualization, this method is invaluable in quantifying and presenting anatomical structures, or assessing the size of pathologies like tumours or blood vessels \cite{iqbal2022g, iqbal2023robust}, an essential aspect in tracking disease evolution or gauging the effectiveness of treatments.

\begin{figure*}[ht]
    \centering
    \includegraphics[scale=0.47]{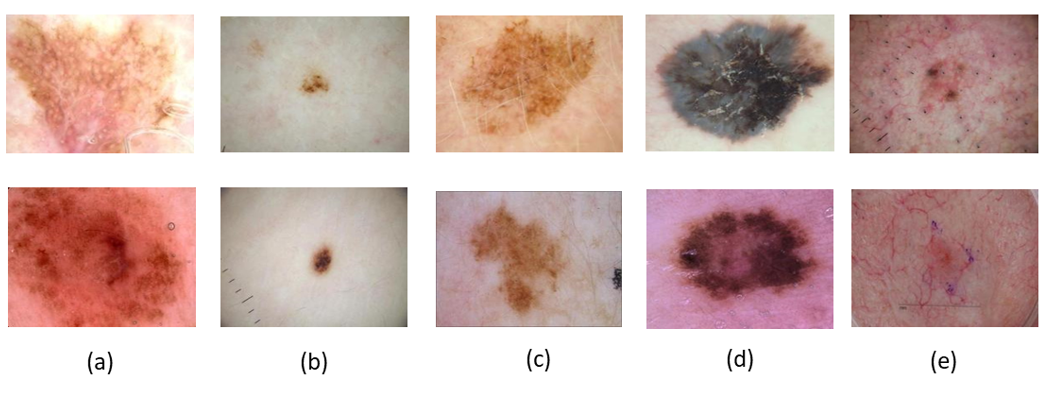}
    \caption{Low inter-class variations between melanoma (1st row) and non-melanoma (2nd row). High intra-class variations in the same class (a-e).}
    \label{Fig: Framework}
    \vspace{0.5cm}
\end{figure*}

CAD systems employ a mix of machine learning and digital image processing techniques to aid dermatologists in clinical screening. Early CAD systems used pre-processing, segmentation of skin lesions, and hand-crafted feature extraction of the segmented lesions for skin lesions classification. 
These methods do not perform well in cases where the lesions have a fuzzy border and the presence of artefacts such as hair and bubbles and ruler marks \cite{oliveira2018computational}. Convolutional neural networks (CNN) have regularly been employed to categorise skin lesions \cite{bi2017automatic}. Although CNN-based skin lesions classification networks have obtained promising results, they are not currently applicable in clinical settings. 
The following factors affect the performance of CAD systems for skin lesions diagnosis: First, the skin lesions are very similar in colour, shape and texture (i.e., low inter-class variation). Second, the high intra-class variations within the same class. The two  challenges are shown in Figure~\ref{Fig: Framework}.

The skin lesions data distribution among different classes is highly imbalanced. The data distribution for the categorization of skin lesions is shown in Table~\ref{tab:datasethamisic19}. The NV class, which has more skin lesion samples, may receive greater attention from CNN models than classes with fewer samples. Most CNN networks fail to generalize on the under-represented classes. A few methods used different approaches (e.g., guided loss function \cite{fernando2021dynamically, gessert2019skin}, auxiliary learning \cite{wei2022dual}) to handle these issues. Due to the aforementioned problems, the categorization of skin lesions remains a difficult procedure even with image-level supervision. 

The proposed method tackles the aforementioned issues by extracting features unique to each class. We observe that each skin lesion class may be represented by a number of feature channels, which aids discrimination of features from various classes. 
We put forward a unique class-wise attention mechanism that determines the importance of features on an image-by-image basis. Multi-scale class-wise feature information is incorporated in a progressive manner. Depth-wise separable convolution is employed in the proposed class-wise attention mechanism to lower the number of trainable parameters.
 
Following is a summary of the proposed framework's main contributions:
\begin{itemize}
\item A novel class-wise attention mechanism is proposed to address the issues of inter-class similarity, intra-class variation and class imbalance.

\item The proposed class-wise attention mechanism is employed progressively to incorporate relevant features from multiple scales.

\item On skin lesions classification benchmarks, the proposed technique obtained state-of-the-art performance in comparison to more than 15 methods from literature including leaderboard winners from ISIC 2019 and HAM10000.
\end{itemize}

The paper's main body is organised as shown below: Section 2 discusses the related works. The architecture of the proposed method is described in Section 3. Extensive experimental results and performance comparisons with earlier methods are presented in Section 4. Section 5 provides the conclusion.

\section{Related Work}
The classification of skin lesions and recent studies using attention mechanisms are briefly reviewed in this section.

\subsection{\textbf{Skin lesions Classification}}
Machine learning (ML) based image recognition and classification have emerged as hot topics in a variety of academic areas, particularly with deep learning. For instance, CNNs have shown potential in the classification of skin lesions \cite{gonzalez2018dermaknet, 9018274}. However, current research using data-driven models has revealed the best performance results across several test sets ever released. These pre-trained models are often used in conjunction with a Transfer Learning (TL) method \cite{ hosny2019classification}.

In practice, image classification demands the employment of interpretations that extract discriminatory features within class variability while preserving informative intra-class variations. Deep neural networks trained by successively using linear and non-linear operations to create hierarchical invariant representations. The majority of image classification problems have general, generalized learnable representations that can be applied to a variety of areas, and they are learnt using different datasets. For the majority of images, data augmentations like translation, rotation, and scale are frequently used when several samples of the same object are included in a database. Utilizing these techniques, the practical implementation of automated skin cancer categorization has substantially improved \cite{gessert2019skin, 8990108, liu2020deep, estava2017dermatologist, khan2021leveraging, khan2022t}. Meanwhile, despite the significant research advancements, a number of variables continue to prevent further improvement in diagnostic performance. The categorization of skin lesions is a difficult process since the images of the lesions typically have poor contrast, fuzzy boundaries, and artefacts including hair, veins, ruler lines, low inter-class variations, substantial intra-class variances, and data imbalance problems. The usage of millions of images in one of the most popular image classification datasets, named ImageNet, demonstrates the requirement for large datasets of images to efficiently train CNN models to learn the properties of the data. \cite{deng2009imagenet}.

The Deep CNN algorithms have produced impressive classification results, gratitude to preliminary training on this extensive dataset for image classification (\cite{estava2017dermatologist, liu2020deep}). Furthermore, data imbalance affects practically all of the publicly available datasets on skin lesions. Because different forms of skin lesions develop at varying rates and are more or less accessible for image capture, samples within different lesion categories exhibit unequal distributions. Additionally, a number of images show low inter-class variances \cite{yu2016automated, yang2019self}. These restrictions results in poor performance.
Improving the Deep CNN architecture from the initial AlexNet \cite{krizhevsky2012imagenet} to a more modern model with greater model parameter ability \cite{radosavovic2020designing} allows for better outcomes on huge image classification tasks. However, it is particularly difficult to enhance performance in small-scale image samples by introducing more parameters, the model could go from a domain with low overfitting risk to one with high overfitting risk \cite{belkin2019reconciling}. Sometimes, Deep CNNs are utilized in studies that are published in the research without taking the new database of images and the additional intra-class or inter-class changes into account. Techniques that lessen the overfitting issue can be used to prevent these issues \cite{estava2017dermatologist, liu2020deep, yu2016automated, HAN20181529, hosny2019classification}. 
These techniques include data augmentations \cite{9025648}, transfer learning-based processes \cite{hosny2019classification}, and multi-target weighted loss functions \cite{fernando2021dynamically}.

\begin{table*}[htbp]
  \centering
  \caption{Data distribution of HAM10000 and ISIC 2019.}
  \adjustbox{width=\textwidth}{
    \begin{tabular}{cccccc}
    \hline
    \multicolumn{1}{l}{\multirow{2}[4]{*}{\textbf{Class Name}}} & \multicolumn{2}{c}{\textbf{HAM10000  }} &       & \multicolumn{2}{c}{\textbf{ISIC2019}} \\
\cmidrule{2-3}\cmidrule{5-6}          & \multicolumn{1}{c}{\textbf{Images}} & \multicolumn{1}{c}{\textbf{Percentage}} &       & \multicolumn{1}{c}{\textbf{Images}} & \multicolumn{1}{c}{\textbf{Percentage}} \\
    \hline
    \multicolumn{1}{l}{MEL} & 1113  & 11.11\% &       & 4522  & 17.85\% \\
    \multicolumn{1}{l}{NV} & 6705  & 66.95\% &       & 12875 & 50.83\% \\
    \multicolumn{1}{l}{BCC} & 514   & 5.10\% &       & 3323  & 13.12\% \\
    \multicolumn{1}{l}{AKIEC} & 327   & 3.30\% &       & 867   & 3.42\% \\
    \multicolumn{1}{l}{BKL} & 1099  & 11\%  &       & 2624  & 10.36\% \\
    \multicolumn{1}{l}{DF} & 115   & 1.20\% &       & 239   & 0.94\% \\
    \multicolumn{1}{l}{VASC} & 142   & 1.40\% &       & 253   & 1\% \\
    \multicolumn{1}{l}{SCC} & -     & -     &       & 628   & 2.48\% \\
    \hline
    Total & 10015 &       &       & 25331 &  \\
    \hline
    \end{tabular}%
    }
  \label{tab:datasethamisic19}%
\end{table*}%
\subsection{\textbf{Deep learning's based attention mechanisms}}
The objective of the deep learning attention mechanism is to emulate the human visual attention system by selecting information from a large amount of data that is most related to the current task. Natural language processing and computer vision have both made considerable strides because of it \cite{ wang2016attention, chen2016neural, xia2021attention}. Numerous attention-based techniques have recently been developed to improve CNN's ability to represent features in image classification tasks. A unique residual attention module was developed by Wang et al. \cite{wang2017residual}. The trainable convolutional layers learn the attention weights, and attention feature maps are produced by fusing the convolutional feature maps with the attention weights. A new squeeze and excitation channel-wise attention network (SENet) was proposed by Hu et al. \cite{hu2018squeeze}. The attention feature maps in each SE block are created by multiplying the input feature maps by the attention vector that was learnt by two successive fully connected (FC) layers. Numerous attention-based methods for classifying skin lesions have been proposed, another One is ARL-CNN \cite{ zhang2019attention}. ARL-CNN can produce attention weights during network training, this reduces the need for processing resources and offers the benefit of avoiding overfitting on small training data sets in contrast to RAN \cite{wang2017residual} and SENet \cite{hu2018squeeze}, which learn attention weights via extra learnable layers. Additionally, Through adaptation, ARL-CNN has a large potential to discriminatively concentrate on skin lesion regions. But, one drawback of ARL-CNN is that it only takes into account spatially focused attention while ignoring crucial channel-wise information. DeMAL-CNN \cite{he2022deep} proposed a mixed attention mechanism that takes into account both spatial and channel-wise attention information while building and integrating it into the architecture of each embedding extraction network. This might help DeMAL-CNN better locate skin lesions for CNN.
 CABNet \cite{he2020cabnet} use global attention and category attention to handle the issue of class imbalance distributions in diabetic retinopathy grading.

\begin{figure*}[ht]
    \centering
    \includegraphics[width=\textwidth]{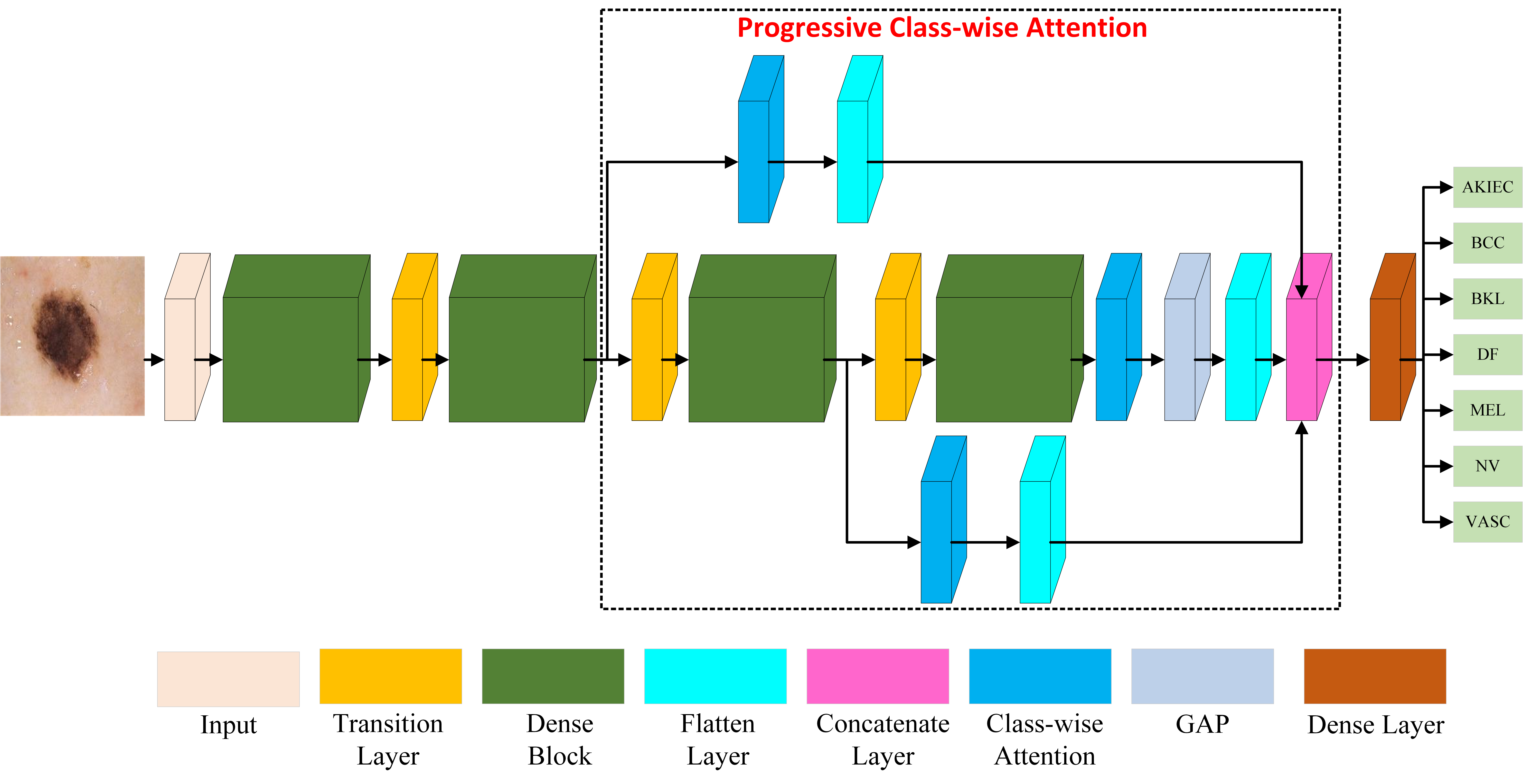}
    \caption{Block level details of the proposed architecture.}
    \label{Frameworkblock}
     \vspace{0.5cm}
\end{figure*}
\begin{figure*}[ht]
    \centering
    \includegraphics[scale=0.65]{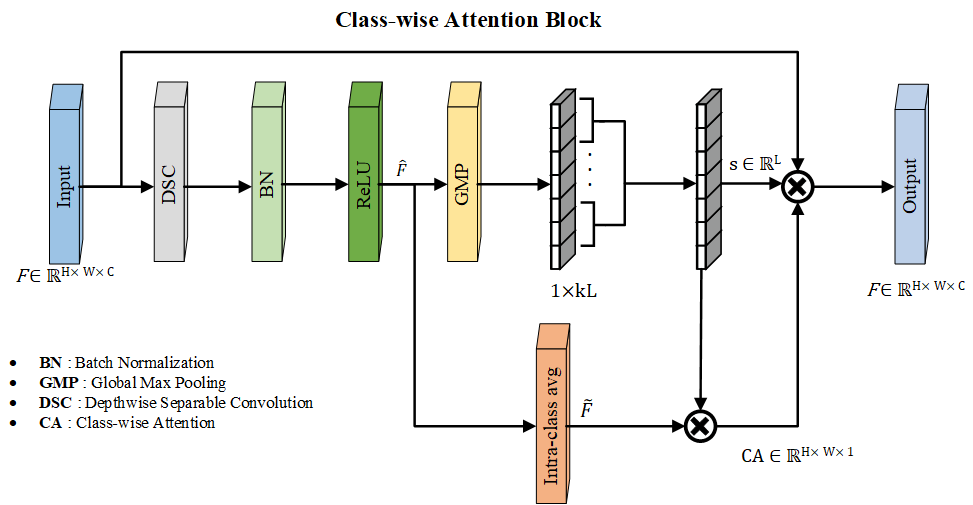}
    \caption{Details of the proposed class-wise attention block.}
    \label{Framework4}
\end{figure*}

\section{Materials and Methods}
In this section, we first go over the benchmark datasets for analyzing skin lesions, then we go over the specifics of the suggested methodology.

\subsection{\textbf{Skin lesions datasets}}

The International Skin Imaging Collaboration (ISIC) obtained and assembled the datasets. The HAM10000 \cite{tschandl2018ham10000} was presented in 2018 containing 10,015 dermoscopic images and having a size of 450 $\times$ 600 pixels. Seven classes make up the HAM10000: nevus (NV), benign keratosis (BKL), basal cell carcinoma (BCC), dermatofibroma (DF), vascular lesion (VASC), actinic keratosis (AKIEC), melanoma (MEL). 
The ISIC 2019 \cite{combalia2019bcn20000, codella2018skin, tschandl2018ham10000} consists of eight classes of dermoscopic images. The classes are nevus (NV), benign keratosis (BKL), basal cell carcinoma (BCC), dermatofibroma (DF), squamos cell carcimoma (SCC), vascular lesion (VASC), actinic keratosis (AKIEC), and melanoma (MEL). The data distribution of each class is present in Table~\ref{tab:datasethamisic19}. 
The ISIC 2019 and HAM10000 are standard benchmark datasets of dermoscopic images for comparative analysis of skin lesion analysis methods. From Table~\ref{tab:datasethamisic19} it can be observed the severity of class imbalance in the above-mentioned datasets. 

\subsection{\textbf{Methodology}} \label{methodology}
Traditional CNN-based approaches do not take into account class-related information, therefore their performance is more adversely affected due to the class imbalance problem in skin lesions classification.
The proposed technique learns the feature maps in a class-specific approach and treats every type of skin lesion equally. 

This is achieved by the proposed class-wise attention technique that associates feature channels with classes equally to reduce channel bias and data redundancy. Results show that class-wise attention mechanisms may effectively address the persistent problems of high inter-class similarity, large intra-class variability, and uneven data distribution in skin lesions classification datasets (Section~\ref{experiments_and_results}).
The details of the proposed network that adopts the DenseNet-121 \cite{huang2017densely} as its backbone is featured in Figure~\ref{Frameworkblock}. The details of the class-wise attention mechanism is presented in Figure~\ref{Framework4}.

\subsubsection{\textbf{Overview of the proposed architecture}}
To extract the feature maps from images of skin lesions, the pre-trained DenseNet-121 is employed as the baseline model. The dense connections in the DenseNet architecture allow maximal transmission of skin lesion characteristics across layers through skip connections. Besides, the dense connections can reuse features and diverse features are produced to cater for the multi-scale skin lesion feature extraction for the classification of skin lesions. DenseNet-121 consists of four successive dense blocks and three transition layers. The dense blocks are composed of 12, 24, 48, and 32 densely linked convolutional layers, respectively. To balance the computational load of the dense layers, the three transition layers are responsible for contracting the feature information coming from the dense blocks, by the use of 2$\times$2 pooling layers. We dropped the last three layers of the baseline model and added more layers. The newly added global average pooling (GAP) and classification output layer have replaced the final three layers of the baseline model. The newly added layers reduce the number of parameters of the baseline model while maintaining the performance measures as well.  

The proposed class-wise attention is applied progressively to the multi-scale features of DenseNet-121 at multiple stages to getting effective attention feature maps. The choice of the backbone stage that is subjected to attention depends upon the high-level semantic information of the features. The proposed architecture employs class-wise attention after the second, third, and fourth dense blocks, respectively.  

Class-wise attention is used to learn various region-wise characteristics for each class. After the class-wise attention mechanism is employed progressively, a global average pooling, and finally a classification layer is used to get the final results.

\begin{table*}[ht]
  \centering
  \caption{Progressive class-wise attention.}
  \adjustbox{width=\textwidth}{
    \begin{tabular}{lccccc}
    \hline
    {\textbf{Methods}} & {\textbf{$A_{cc}$}} &{\textbf{$P_{r}$}} & {\textbf{$S_{n}$}} & {\textbf{$F_{1}$}} & {\textbf{$AUC$}} \\
    \hline
    baseline (DenseNet-121) & {0.889} & {0.900} &	{0.888} &	{0.887} &	{0.993}\\
    baseline+class-wise attention after 4th dense block
    & {0.958} &{0.963} & {0.957} & {0.957} & {0.997} \\
    baseline+class-wise attention at progressive positions
    & {0.973} &{0.973} & {0.973} & {0.972} & {0.998} \\
    \hline
    \end{tabular}%
    }
  \label{tab:addlabelresultC}%
\end{table*}%

\subsubsection{\textbf{Class-wise attention mechanism}}
To improve the fine-grained classification performance on the specified classes, the class-wise attention mechanism presented in Figure~\ref{Framework4} is proposed. It has the ability to distinguish skin lesions on class-based discriminating characteristics. 

Let the input tensor from a dense block is represented as $\mathcal{F}\in \mathbb{R}^{H\times W\times C} = \{F_1,F_2,\cdots, F_C\}$, where the number of channels, height and width are represented by C, H, and W, respectively. The tensor $\mathcal{F}$ is subjected to a depth-wise separable convolution layer followed by a batch normalization layer and a ReLU activation layer to obtain the tensor $\mathcal{\hat{F}} \in \mathbb{R}^{H \times W \times kL} = \{\hat{F}_1,\hat{F}_2,\cdots,\hat{F}_{kL}\}$, mathematically represented as: 

\begin{equation}\label{}
\mathcal{\hat{F}} = \text{ReLU}(\mu(\mathcal{O}^{3\times3}(\mathcal{F}))),
\end{equation}
where $k$ is the number of feature channels required to find feature maps that can distinguish between each class,
$L$ is the number of classes, $\mu$ is batch normalization and $\mathcal{O}$ is a depth-wise separable convolution layer.
A 3$\times$3 depth-wise separable convolution operation that forms the foundation of our proposed attention technique helps the proposed network to pay attention to specific channels for identifying a particular class.  
Given the tensor $\mathcal{\hat{F}}$, the score for each class is computed by grouping the channels per class according to $k$ as follows:
\begin{equation}\label{}
s_i = \frac{1}{k} \sum_{j=1} ^{k} \delta(\mathcal{\hat{F}}_{i, j}), 
i \in {1,2,\cdots,L},
\end{equation}

where $\delta$ stands for global max pooling. The class score vector $\mathbf{s} = [s_1, s_2,\cdots,s_L]$ signifies the importance of each class for a particular sample.

 The class-wise semantic feature map $\mathcal{\tilde{F}} \in \mathbb{R}^{H \times W \times L}$ is computed by averaging the per class maps as follows
 \begin{equation}\label{}
 \mathcal{\tilde{F}} = \frac{1}{k} \sum_{j=1} ^{k} \mathcal{\hat{F}}_{i,j}.
 i\in {1,2,\cdots,L},   
\end{equation} 

The class-level attention map $CA$ is obtained as per equation~\ref{CA}
\begin{equation}\label{CA}
CA = \frac {1}{L} \sum_{i=1}^ {L} \mathbf{s} \mathcal{\tilde{F}}. 
\end{equation}  
The discriminative regions that are informative for skin lesions classification are highlighted in CA. Finally, equation~\ref{FCA} applies the class-wise attention to the input tensor $\mathcal{F}$
\begin{equation}\label{FCA}
\mathcal{F}_{CA} = \mathcal{F}\otimes CA,  
\end{equation}
where the final output feature tensor after class-wise attention is $\mathcal{F}_{CA}$ and $\otimes$ stands for element wise multiplication.

\subsubsection{\textbf{Loss functions}}

The focal loss \cite{Linetal2020} is employed for multi-class classification as described in Eq.~\ref{equationfocalloss}.
\begin{equation}\label{equationfocalloss}
\mathcal{L}_{\text{focal}} = -\sum_{i=1} ^{N} (1-\hat{p}_y)^\gamma log (\hat{p}_y),
\end{equation}
where $\hat{p}$ is the probability estimate for class $y$. In our approach, $gamma$ serves as a weighting factor for the challenging samples and is empirically determined to be 2.

\section{Experiments and Results}\label{experiments_and_results}
This section reports the outcomes of the proposed approach on benchmark datasets relevant to skin lesions classification, including the ablation study and qualitative and quantitative findings. For interpretability of the learned solution, Grad-CAM \cite{selvaraju2017grad} is used to obtain the attention maps of the proposed framework.

\begin{table*}[htbp]
  \centering
  \caption{Average results of the proposed method on HAM10000 over five training runs along with the standard deviation.}
  \adjustbox{ width=\textwidth}{
    \begin{tabular} {lcccccc}
    \hline
   \textbf{Method} & Runs  & {\textbf{$A_{cc}$}} & {\textbf{$P_{r}$}} & {\textbf{$S_{n}$}} & {\textbf{$F_{1}$}} & {\textbf{${AUC}$}}  \\
   
    \hline
    {PCA} & First & 0.9768 & 0.9777 & 0.9768 & 0.9763 & 0.9981 \\
    & Second & 0.9734 & 0.9739 & 0.9734 & 0.9728 & 0.9976 \\
    & Third & 0.9778 & 0.9781 & 0.9778 & 0.9774 & 0.9982 \\
    & Fourth & 0.9720 & 0.9729 & 0.9719 & 0.9713 & 0.9977 \\
    & Fifth & 0.9713 & 0.9712 & 0.9713 & 0.9709 & 0.9977 \\
    \hline
    \textbf{Proposed (ours)} & &{0.9743 $\pm$ 0.0029} & \textbf{0.9748 $\pm$ 0.0030} & \textbf{0.9742 $\pm$ 0.0029}& \textbf{0.9737 $\pm$ 0.0030} & \textbf{0.9978 $\pm$ 0.0027} \\
    \hline
    \end{tabular}%
    }
  \label{tab:addlabel5turn}%
\end{table*}
 
\subsection{\textbf{Implementation details}}

This section outlines the proposed approach implementation details using benchmark datasets for skin lesions classification. Based on prior works \cite{9264664, wei2022dual, kassem2020skin}, we up-sampled the datasets by duplicating the images of minority classes to equal the images in the majority class (i.e., NV class). Afterwards, the datasets were divided into three sections: 20\% for test, 20\% for validation, and 60\% for training. 

 To overcome the issue of less data, we performed data augmentation to enhance data diversity including data rotation from 0 to 180 degrees, width and height shift, zoom in and out, and horizontal and vertical flip. Before augmentation, the datasets were resized to 224$\times$224 pixel resolution. We employed focal loss with Nadam optimizer for the period of 40 iterations with early stopping. Depending on validation results, the learning rate gradually decreased after five epochs from its original value of 0.001 with a factor of 0.25. In order to build the proposed framework, Keras and TensorFlow are used as the backend. The NVIDIA K80 GPU is used to train all model variations.
 
\subsection{\textbf{Evaluation metrics}}
Every multi-class classification method is evaluated by the standard performance metrics \cite{grandini2020metrics}. Therefore, the following metrics are used in this study to assess the effectiveness of the methods: accuracy, F1 score, precision, sensitivity, specificity, and AUC (Area Under Curve). The value of performance measures is determined using the true positive (TP), false positive (FP), true negative (TN), and false negative (FN) predictions.\\
Accuracy: counts the number of accurate predictions made relative to all other predictions, it is frequently used as the basic measure for model evaluation \cite{grandini2020metrics}. It is calculated as
\begin{equation}\label{equation7}
A_{cc} = \frac{{TP+TN}}{{TP+TN+FP+FN}}.
\end{equation}
Precision: It is defined as the ratio of correctly classified positive samples (TP) to all positively classified samples (either correctly or incorrectly) \cite{grandini2020metrics}. Eq.~\ref{equation8} is used to calculate the precision of a method. Precision is important for medical image diagnosis as it highlights false positive cases.
\begin{equation}\label{equation8}
P_r = \frac{{TP}}{{TP + FP}}
\end{equation}
Sensitivity: The percentage of positive samples that were correctly classified as positive relative to all positive samples. The model's sensitivity measures how effectively it can detect positive samples \cite{grandini2020metrics}. Eq.~\ref{equation9} is used to calculate the sensitivity value. For medical image diagnosis, sensitivity is also an important measure. For better diagnosis, the value of false negatives in a CAD system should be minimum.
\begin{equation}\label{equation9}
S_n = \frac{{TP}}{{TP + FN}}
\end{equation}
F1 Score: Precision and recall are harmonically summed to produce the F1 score. It is a helpful measurement for datasets with imbalance distribution \cite{grandini2020metrics}. Eq.~\ref{equation10} is used to calculate the F1 score 
\begin{equation}\label{equation10}
F_1 = 2*\frac{{Precision\times\ Recall}}{{Precision + Recall}}.
\end{equation}
Specificity: The percentage of real negatives that the model successfully predicted \cite{grandini2020metrics}. Eq.~\ref{equation11} explains the specificity 
\begin{equation}\label{equation11}
S_p = \frac{{TN}}{{TN + FP}}.
\end{equation}
Area Under Curve (AUC) refers to the evaluation result under the Receiver Operating Characteristic (ROC) curve. The AUC is the performance indicator for classification models at different threshold levels. The model's ability to distinguish across classes are evaluated with the help of these measures.

\subsection{\textbf{Ablation study}}

We conducted an ablation study using the proposed methodology on the HAM10000 dataset to understand the impacts of each component of the proposed methodology, in particular the attention mechanism. We compare the baseline method, i.e., DenseNet-121 with two proposed class-wise attention variants. For the first variant, the proposed class-wise attention mechanism is employed after the baseline model's fourth dense block. The second variant includes the class-wise attention mechanism at successive stages of the baseline model (i.e., after the 2nd, 3rd and 4th dense blocks), which we call progressive class-wise attention. The findings of the ablation study are displayed in Table~\ref{tab:addlabelresultC}.
To observe the confidence of the predictions of the progressive class-wise attention, we computed its five distinct runs on the HAM10000 dataset. The average results along with the standard deviation are presented in Table~\ref{tab:addlabel5turn}, which demonstrate the confidence of the proposed methodology on its predictions.
\subsection{\textbf{Results comparison with other attention mechanisms}}
Now we compare the performance of our approach with other state-of-the-art attention techniques. For a fair comparison, the attention mechanisms from the literature are employed at the same position in the DensNet architecture as the proposed class-wise attention. Table \ref{tab:addlabel1} compares the outcomes of the attention technique for the HAM10000 dataset. The results demonstrate that the suggested attention performs effectively on the task of classifying skin lesions as compared with its counterparts. This improvement in performance is achieved while consuming fewer parameters as compared with the alternatives. 

CBAM is a benchmark attention mechanism and has been employed in various methods targeted to diverse applications \cite{alirezazadeh2022improving, yu2021application, li2022eca}. Table~\ref{tab:addlabel1} demonstrates how our proposed attention mechanism outperforms CBAM with fewer computational parameters in terms of accuracy, precision, recall, F1 score, and AUC. In comparison with CABNet our proposed method achieved competitive performance with a 1.1\% improvement in precision with almost 2 million fewer parameters. 
\begin{table*}[htbp]
  \centering
  \caption{Proposed class-wise attention comparison with other attention mechanisms.}
     \adjustbox{width=\textwidth}{
    \begin{tabular}{lcccccc}
    \hline
    {\textbf{Methods}} & {\textbf{$A_{cc}$}} & {\textbf{$P_{r}$}} & {\textbf{$S_{n}$}} & {\textbf{$F_{1}$}} & {\textbf{${AUC}$}} & {\textbf{$Parameters$}}
    \\
    \hline
    DenseNet-121 (Baseline) & 0.889 & 0.900 & 0.888 & 0.887 & 0.993 & 6.961 M\\
    
    DenseNet-121+CBAM \cite{woo2018cbam} & 0.928 & 0.929 & 0.928 & 0.926 & 0.994 & 7.225 M\\
    {DenseNet-121+Soft Attention \cite{datta2021soft}} & 0.951 & 0.951 & 0.950 & 0.950 & 0.996 & 7.329 M\\
    {DenseNet-121+CABNet \cite{he2020cabnet}} & 0.953 & 0.952 & 0.953 & 0.952 & 0.997 & 9.176 M\\
    \textbf{DenseNet-121+Class-wise Attention (ours)} & \textbf{0.958} & \textbf{0.963} & \textbf{0.957} & \textbf{0.957} & \textbf{0.997} & \textbf{7.085 M}\\
    \hline
    \end{tabular}%
    }
  \label{tab:addlabel1}%
\end{table*}%

\subsection{\textbf{Comparison of the HAM10000 dataset's results}}
The experimental findings of the suggested strategy for categorising exceedingly complex skin lesion types of the HAM10000 dataset are presented in this section. There are seven classes in this dataset that belong to dermoscopy images of resolution 450 $\times$ 600. 
In Table~\ref{tab:addlabelresultC12}, the comparative analysis of the proposed method on the HAM10000 dataset is presented. We compared the proposed approach findings with existing approaches from literature and the top three positions on the ISIC archive leader board.

Meta-Optima's \cite{nozdryn2018ensembling} secured first place in the HAM10000 skin cancer categorization competition. Stacked ensemble CNN models are the foundation of the winner technique. The training set has been increased by the number of open-source skin cancer datasets, and several data augmentation techniques have been used. To address the issue of class imbalance, they have employed mini-batch, external data, and the inverse frequency, which selects the weight of the loss function. Our proposed method outperforms the ISIC leader-board winner with performance improvements of 1.7\%, 18\%, 16.9\%, 18.3\% and 1.5\% in terms of accuracy, precision, sensitivity, F1 score and AUC, respectively.
\begin{table*}[ht]
  \centering
  \caption{The results comparison on the HAM10000 data set.}
  \adjustbox{max width=\textwidth}{
    \begin{tabular}{lccccc}
    \hline
    {\textbf{Methods}} & {\textbf{$A_{cc}$}} &{\textbf{$P_{r}$}} & {\textbf{$S_{n}$}} & {\textbf{$F_{1}$}} & {\textbf{$AUC$}} \\
    \hline
    Winner ISIC 2018 \cite{nozdryn2018ensembling} & 0.958 & 0.826 & 0.833 & 0.823 & 0.983\\
    Runner-Up ISIC 2018 \cite{gessert2018skin} & 0.960 & 0.838 & 0.835 & 0.831 & 0.982\\
    Rank3 ISIC 2018 \cite{zhuang2018skin} & 0.954 & 0.794 & 0.830 & 0.805 & 0.980 \\
    \hline
    Dual Attention \cite{wei2022dual} & 0.896 & - & 0.817 & - & 0.983\\
    FTN \cite{he2022fully} & 0.927 & 0.621 & 0.857& - & 0.973 \\ 
    VIT Model \cite{xin2022improved} & 0.941 & 0.942 & - & 0.941& 0.987 \\
    MSM-CNN \cite{mahbod2020transfer} & 0.963 & 0.913 & - & - & 0.981 \\ 
    CAD \cite{bakkouri2020computer} & \textbf{0.981} & 0.935 & 0.934 & 0.934 & 0.965 \\
     FcResNet-TL \cite{razzak2020skin} & 0.961 & 0.960 & 0.964 & 0.961 & 0.984 \\
    Unit-vise \cite{9264664} & \textbf{0.981} & 0.911 & 0.972 & 0.973 & 0.984\\
    \hline
     \textbf{Proposed (ours)} & {0.974} & \textbf{0.975} & \textbf{0.974}& \textbf{0.974} & \textbf{0.998} \\
      \hline
    \end{tabular}%
    }
  \label{tab:addlabelresultC12}%
\end{table*}%

Our proposed approach exceeded current state-of-the-art methods on the majority of performance metrics. The CAD \cite{bakkouri2020computer} approach achieved the highest accuracy on the HAM10000 dataset. The CAD method used transfer learning, multi-layer feature fusion network and up-sampling and down-sampling approaches. The dual attention \cite{wei2022dual} method is used to channel spatial attention with auxiliary learning. To address the issue of imbalance the authors have also used an up-sampling approach. The FTN \cite{he2022fully} and VIT Model \cite{xin2022improved} are transformer-based methods. The VIT model uses label shuffling and patch-based training to handle class imbalance. The FTN method also used transfer learning for skin lesions classification. The MSM-CNN \cite{mahbod2020transfer} employed three ensembles of learning-based CNN models to classify skin lesions. The authors employed external data and different sizes of cropped images for classification. FcResNet-TL \cite{razzak2020skin} and Unit-vise \cite{9264664} performed up-sampling and different data augmentations to enhance the data set. 

\begin{figure*}[ht]
    \centering
    \includegraphics[scale=.45]{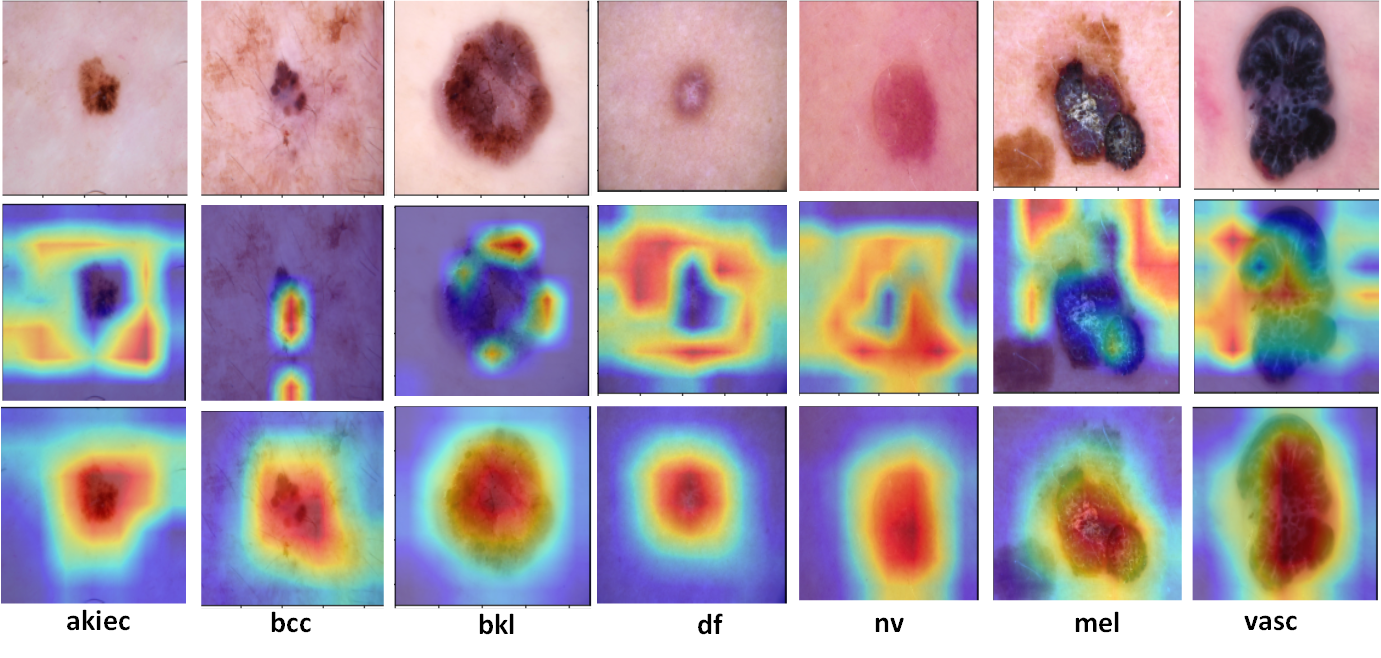}
    \caption{Visualization results of each class on HAM10000 dataset. Original images are displayed in the first row, the second row shows the attention maps of the baseline model and the third row shows the attention maps of the proposed method.}
    \label{Framework5}
\end{figure*}

To demonstrate the suggested model's capacity for prediction and the interpretability of the attention processes, we utilized Grad-CAM \cite{selvaraju2017grad}. The generated attention maps can assist dermatologists in analysing the lesion area for a particular class. 
In the first row of Figure~\ref{Framework5}, we present images from the HAM10000 data set that feature different classes. The second and third rows show the attention maps of the baseline and the proposed method, respectively. 
As seen in Figure~\ref{Framework5}, the model without attention mechanism emphasises on unwanted regions, hence missing important class-related feature information. Contrarily, the proposed framework concentrates on the important lesion areas that are necessary for robust skin cancer classification. 
Overall, the class activation maps demonstrate that the proposed approach extracts more precise lesion areas for diagnosis purposes.

\subsection{\textbf{Results comparison on the ISIC 2019 dataset}}

This section presents the experimental results of the proposed approach on ISIC 2019 dataset.
 
The findings of the proposed technique are compared with those of other state-of-the-art techniques in Table~\ref{tab:addlabelresultISIC19}.
The winner of the ISIC 2019 leaderboard used an ensemble approach and balancing loss to perform the classification of skin lesions. They also used different image resolutions, cropping and data augmentation approaches to enhance the data. Our proposed method surpassed in performance in comparison to the winner of the ISIC 2019 with improvements in terms of accuracy (0.1\%), precision (55\%), sensitivity (66\%), F1 score (63.6\%), specificity (1.8\%) and AUC (5.6\%).     

\begin{table*}[ht]
  \centering
  \caption{The results comparison on the ISIC 2019 data set.}
  \adjustbox{max width=\textwidth}{
    \begin{tabular}{lcccccc}
    \hline
    {\textbf{Methods}} & {\textbf{$A_{cc}$}} &{\textbf{$P_{r}$}} & {\textbf{$S_{n}$}} & {\textbf{$F_{1}$}} & {\textbf{$S_{p}$}} & {\textbf{$AUC$}} \\
    \hline
    Winner ISIC 2019 \cite{gessert2019skin} & 0.940 & 0.609 & 0.571 & 0.578 & 0.975 & 0.941\\
    Runner-Up ISIC 2019 \cite{zhou2019multi} & 0.932 & 0.515 & 0.661 & 0.567 & 0.951 & 0.807\\
    \hline
    Dual Attention \cite{wei2022dual} & 0.890 & - & 0.835 & - & 0.979 &0.976\\
    Modified GoogleNet \cite{kassem2020skin} & 0.949  & 0.804 & 0.798 & 0.801 & 0.970 & -\\
    Fine-tuned DenseNet201 \cite{benyahia2022multi} & 0.923 & 0.852 & 0.928 & 0.870 & 0.964 & -\\
    Optimized DenseNet-201 \cite{villa2022optimized} & 0.930 & 0.940 & 0.930 & 0.930 & 0.930 & 0.965\\
    \hline
     \textbf{Proposed (ours)} & \textbf{0.949} & \textbf{0.949} & \textbf{0.948}& \textbf{0.947} & \textbf{0.993} & \textbf{0.994} \\
      \hline
    \end{tabular}%
    }
  \label{tab:addlabelresultISIC19}%
\end{table*}%

The modified GoogleNet \cite{kassem2020skin} achieved the results using transfer learning and by adding more filters in the architecture with up-sampling and down-sampling approaches. They achieved the best results by the down-sampling approach. Our proposed method achieves higher results in terms of precision (18\%), sensitivity (18.8\%), F1 score (18.2\%) and specificity (2.4\%). The author of Optimized DenseNet-201 \cite{villa2022optimized} also used a balanced dataset for the evaluation of results with the modification in DenseNet architecture. The Fine-tuned DenseNet201 \cite{benyahia2022multi} method also uses up-sampling and down-sampling approaches with some image pre-processing techniques for classification tasks.\\
Table~\ref{tab:addlabelresultISIC19} shows that the outcomes of the proposed technique on the ISIC2019 dataset are excellent. The proposed technique outperformed the comparison techniques in every performance measure.

\begin{figure*}[ht]
    \centering
    \includegraphics[scale=.40]{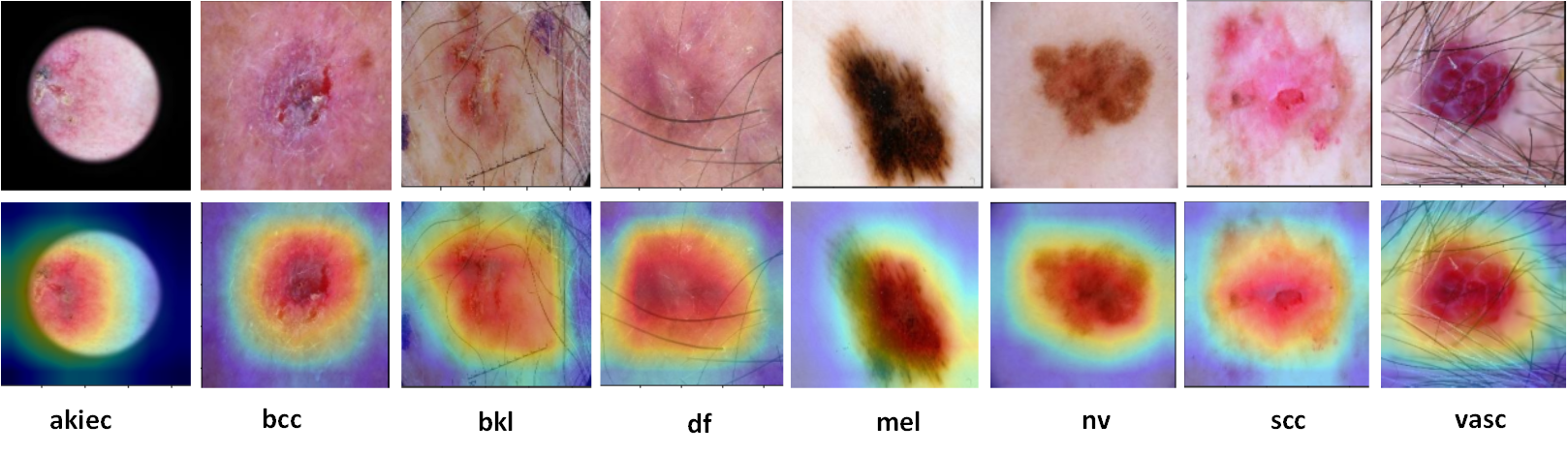}
    \caption{Visualisation results of the ISIC 2019 dataset. The first row contains the original images, while the second row displays the outcomes of the proposed approach.}
    \label{Frameworkisic19}
\end{figure*}
In Figure~\ref{Frameworkisic19}, we present the attention maps of the ISIC 2019 dataset. In Figure~\ref{Frameworkisic19}, the first row represent the original images and the second row shows the attention maps of the proposed model.
The proposed framework classification findings are well predictable since the activation maps concentrate clearly on suspicious lesion regions, which will be helpful for clinical diagnosis. 

The proposed strategy is thoroughly discussed in terms of numerical performances and visual outcomes. The proposed method achieved state-of-the-art results on both datasets owing to its class-wise attention mechanism that learns the features in a category-wise approach and treats every type of skin lesion in an identical manner. Since traditional CNN just stacks all the feature maps together without making any distinctions about classes, the information about the various categories is difficult to distinguish for correct classification.
In order to lessen channel bias and expand the gap between distinct skin lesion classes, the proposed progressive class-wise attention assigns each type of skin lesion a certain number of feature channels and ensures that each skin lesion type has an equal number of feature channels. From experimental results and attention maps, it can be seen that the proposed method produced state-of-the-art classification performance for skin lesions.

\section{Conclusion}
In this study, we present a progressive class-wise attention technique that is end-to-end trainable and has a strong generalisation of unseen data for the classification of skin lesions.
The issue of high inter-class similarity, high intra-class variation, and class imbalance of skin lesions classification is addressed by the proposed technique. The proposed progressive class-wise attention focused on the discriminative areas of dermoscopic images to improve skin lesions classification performance. We employ Grad-CAM to gain insights into the proposed method's learned solutions and to aid clinical experts. Extensive analyses on two benchmark datasets for the classification of skin lesions are used to demonstrate the efficacy of the proposed approach. The efficiency of the proposed method can be enhanced through the incorporation of clinical data (e.g., age, sex, and body part information).

\bibliographystyle{IEEEtran}
\bibliography{references}

\begin{thebibliography}{10}
\providecommand{\url}[1]{#1}
\csname url@samestyle\endcsname
\providecommand{\newblock}{\relax}
\providecommand{\bibinfo}[2]{#2}
\providecommand{\BIBentrySTDinterwordspacing}{\spaceskip=0pt\relax}
\providecommand{\BIBentryALTinterwordstretchfactor}{4}
\providecommand{\BIBentryALTinterwordspacing}{\spaceskip=\fontdimen2\font plus
\BIBentryALTinterwordstretchfactor\fontdimen3\font minus
  \fontdimen4\font\relax}
\providecommand{\BIBforeignlanguage}[2]{{%
\expandafter\ifx\csname l@#1\endcsname\relax
\typeout{** WARNING: IEEEtran.bst: No hyphenation pattern has been}%
\typeout{** loaded for the language `#1'. Using the pattern for}%
\typeout{** the default language instead.}%
\else
\language=\csname l@#1\endcsname
\fi
#2}}
\providecommand{\BIBdecl}{\relax}
\BIBdecl

\bibitem{razzak2020skin}
I.~Razzak, G.~Shoukat, S.~Naz, and T.~M. Khan, ``Skin lesion analysis toward
  accurate detection of melanoma using multistage fully connected residual
  network,'' in \emph{2020 International Joint Conference on Neural Networks
  (IJCNN)}.\hskip 1em plus 0.5em minus 0.4em\relax IEEE, 2020, pp. 1--8.

\bibitem{9264664}
I.~Razzak and S.~Naz, ``Unit-vise: Deep shallow unit-vise residual neural
  networks with transition layer for expert level skin cancer classification,''
  \emph{IEEE/ACM Transactions on Computational Biology and Bioinformatics},
  vol.~19, no.~02, pp. 1225--1234, mar 2022.

\bibitem{siegel2021cancer}
L.~Siegel~Rebecca, D.~Miller~Kimberly, E.~Fuchs~Hannah, and A.~Jemal, ``Cancer
  statistics, 2021,'' \emph{CA Cancer J Clin}, vol.~71, no.~1, pp. 7--33, 2021.

\bibitem{saeed2021skin}
J.~Saeed and S.~Zeebaree, ``Skin lesion classification based on deep
  convolutional neural networks architectures,'' \emph{Journal of Applied
  Science and Technology Trends}, vol.~2, no.~01, pp. 41--51, 2021.

\bibitem{okur2018survey}
E.~Okur and M.~Turkan, ``A survey on automated melanoma detection,''
  \emph{Engineering Applications of Artificial Intelligence}, vol.~73, pp.
  50--67, 2018.

\bibitem{khan2019classification}
M.~Q. Khan, A.~Hussain, S.~U. Rehman, U.~Khan, M.~Maqsood, K.~Mehmood, and
  M.~A. Khan, ``Classification of melanoma and nevus in digital images for
  diagnosis of skin cancer,'' \emph{IEEE Access}, vol.~7, pp. 90\,132--90\,144,
  2019.

\bibitem{dildar2021skin}
M.~Dildar, S.~Akram, M.~Irfan, H.~U. Khan, M.~Ramzan, A.~R. Mahmood, S.~A.
  Alsaiari, A.~H.~M. Saeed, M.~O. Alraddadi, and M.~H. Mahnashi, ``Skin cancer
  detection: a review using deep learning techniques,'' \emph{International
  journal of environmental research and public health}, vol.~18, no.~10, p.
  5479, 2021.

\bibitem{healsmith1994evaluation}
M.~Healsmith, J.~Bourke, J.~Osborne, and R.~Graham-Brown, ``An evaluation of
  the revised seven-point checklist for the early diagnosis of cutaneous
  malignant melanoma,'' \emph{British Journal of Dermatology}, vol. 130, no.~1,
  pp. 48--50, 1994.

\bibitem{mahbod2020transfer}
A.~Mahbod, G.~Schaefer, C.~Wang, G.~Dorffner, R.~Ecker, and I.~Ellinger,
  ``Transfer learning using a multi-scale and multi-network ensemble for skin
  lesion classification,'' \emph{Computer methods and programs in biomedicine},
  vol. 193, p. 105475, 2020.

\bibitem{khawaja2019improved}
A.~Khawaja, T.~M. Khan, K.~Naveed, S.~S. Naqvi, N.~U. Rehman, and S.~J. Nawaz,
  ``An improved retinal vessel segmentation framework using frangi filter
  coupled with the probabilistic patch based denoiser,'' \emph{IEEE Access},
  vol.~7, pp. 164\,344--164\,361, 2019.

\bibitem{khawaja2019multi}
A.~Khawaja, T.~M. Khan, M.~A. Khan, and J.~Nawaz, ``A multi-scale directional
  line detector for retinal vessel segmentation,'' \emph{Sensors}, vol.~19,
  no.~22, 2019.

\bibitem{khan2021residual}
T.~M. Khan, A.~Robles-Kelly, S.~S. Naqvi, and A.~Muhammad, ``Residual
  multiscale full convolutional network (rm-fcn) for high resolution semantic
  segmentation of retinal vasculature,'' in \emph{Structural, Syntactic, and
  Statistical Pattern Recognition: Joint IAPR International Workshops, S+ SSPR
  2020, Padua, Italy, January 21--22, 2021, Proceedings}.\hskip 1em plus 0.5em
  minus 0.4em\relax Springer Nature, 2021, p. 324.

\bibitem{khan2022width}
T.~M. Khan, M.~A. Khan, N.~U. Rehman, K.~Naveed, I.~U. Afridi, S.~S. Naqvi, and
  I.~Raazak, ``Width-wise vessel bifurcation for improved retinal vessel
  segmentation,'' \emph{Biomedical Signal Processing and Control}, vol.~71, p.
  103169, 2022.

\bibitem{khan2023neural}
T.~M. Khan, S.~S. Naqvi, A.~Robles-Kelly, and E.~Meijering, ``Neural network
  compression by joint sparsity promotion and redundancy reduction,'' in
  \emph{Neural Information Processing: 29th International Conference, ICONIP
  2022, Virtual Event, November 22--26, 2022, Proceedings, Part I}.\hskip 1em
  plus 0.5em minus 0.4em\relax Springer International Publishing Cham, 2023,
  pp. 612--623.

\bibitem{iqbal2022recent}
S.~Iqbal, T.~M. Khan, K.~Naveed, S.~S. Naqvi, and S.~J. Nawaz, ``Recent trends
  and advances in fundus image analysis: A review,'' \emph{Computers in Biology
  and Medicine}, p. 106277, 2022.

\bibitem{khan2023simple}
T.~M. Khan, M.~Arsalan, I.~Razzak, and E.~Meijering, ``Simple and robust
  depth-wise cascaded network for polyp segmentation,'' \emph{Engineering
  Applications of Artificial Intelligence}, vol. 121, p. 106023, 2023.

\bibitem{naqvi2023glan}
S.~S. Naqvi, Z.~A. Langah, H.~A. Khan, M.~I. Khan, T.~Bashir, M.~Razzak, and
  T.~M. Khan, ``Glan: Gan assisted lightweight attention network for biomedical
  imaging based diagnostics,'' \emph{Cognitive Computation}, vol.~15, no.~3,
  pp. 932--942, 2023.

\bibitem{khan2023retinal}
T.~M. Khan, S.~S. Naqvi, A.~Robles-Kelly, and I.~Razzak, ``Retinal vessel
  segmentation via a multi-resolution contextual network and adversarial
  learning,'' \emph{Neural Networks}, 2023.

\bibitem{iqbal2022g}
S.~Iqbal, S.~S. Naqvi, H.~A. Khan, A.~Saadat, and T.~M. Khan, ``G-net light: A
  lightweight modified google net for retinal vessel segmentation,'' in
  \emph{Photonics}, vol.~9, no.~12.\hskip 1em plus 0.5em minus 0.4em\relax
  MDPI, 2022, p. 923.

\bibitem{iqbal2023robust}
S.~Iqbal, K.~Naveed, S.~S. Naqvi, A.~Naveed, and T.~M. Khan, ``Robust retinal
  blood vessel segmentation using a patch-based statistical adaptive
  multi-scale line detector,'' \emph{Digital Signal Processing}, p. 104075,
  2023.

\bibitem{oliveira2018computational}
R.~B. Oliveira, J.~P. Papa, A.~S. Pereira, and J.~M.~R. Tavares,
  ``Computational methods for pigmented skin lesion classification in images:
  review and future trends,'' \emph{Neural Computing and Applications},
  vol.~29, no.~3, pp. 613--636, 2018.

\bibitem{bi2017automatic}
L.~Bi, J.~Kim, E.~Ahn, and D.~Feng, ``Automatic skin lesion analysis using
  large-scale dermoscopy images and deep residual networks,'' \emph{arXiv
  preprint arXiv:1703.04197}, 2017.

\bibitem{fernando2021dynamically}
K.~R.~M. Fernando and C.~P. Tsokos, ``Dynamically weighted balanced loss: class
  imbalanced learning and confidence calibration of deep neural networks,''
  \emph{IEEE Transactions on Neural Networks and Learning Systems}, 2021.

\bibitem{gessert2019skin}
N.~Gessert, T.~Sentker, F.~Madesta, R.~Schmitz, H.~Kniep, I.~Baltruschat,
  R.~Werner, and A.~Schlaefer, ``Skin lesion classification using cnns with
  patch-based attention and diagnosis-guided loss weighting,'' \emph{IEEE
  Transactions on Biomedical Engineering}, vol.~67, no.~2, pp. 495--503, 2019.

\bibitem{wei2022dual}
Z.~Wei, Q.~Li, and H.~Song, ``Dual attention based network for skin lesion
  classification with auxiliary learning,'' \emph{Biomedical Signal Processing
  and Control}, vol.~74, p. 103549, 2022.

\bibitem{gonzalez2018dermaknet}
I.~Gonzalez-Diaz, ``Dermaknet: Incorporating the knowledge of dermatologists to
  convolutional neural networks for skin lesion diagnosis,'' \emph{IEEE journal
  of biomedical and health informatics}, vol.~23, no.~2, pp. 547--559, 2018.

\bibitem{9018274}
P.~Tang, Q.~Liang, X.~Yan, S.~Xiang, and D.~Zhang, ``Gp-cnn-dtel: Global-part
  cnn model with data-transformed ensemble learning for skin lesion
  classification,'' \emph{IEEE Journal of Biomedical and Health Informatics},
  vol.~24, no.~10, pp. 2870--2882, 2020.

\bibitem{hosny2019classification}
K.~M. Hosny, M.~A. Kassem, and M.~M. Foaud, ``Classification of skin lesions
  using transfer learning and augmentation with alex-net,'' \emph{PloS one},
  vol.~14, no.~5, p. e0217293, 2019.

\bibitem{8990108}
Y.~Xie, J.~Zhang, Y.~Xia, and C.~Shen, ``A mutual bootstrapping model for
  automated skin lesion segmentation and classification,'' \emph{IEEE
  Transactions on Medical Imaging}, vol.~39, no.~7, pp. 2482--2493, 2020.

\bibitem{liu2020deep}
Y.~Liu, A.~Jain, C.~Eng, D.~H. Way, K.~Lee, P.~Bui, K.~Kanada,
  G.~de~Oliveira~Marinho, J.~Gallegos, S.~Gabriele \emph{et~al.}, ``A deep
  learning system for differential diagnosis of skin diseases,'' \emph{Nature
  medicine}, vol.~26, no.~6, pp. 900--908, 2020.

\bibitem{estava2017dermatologist}
A.~Estava, B.~Kuprel, R.~Novoa, J.~Ko, S.~Swetter, H.~Blau, and S.~Thrun,
  ``Dermatologist level classification of skin cancer with deep neural networks
  [j],'' \emph{Nature}, vol. 542, no. 7639, pp. 115--118, 2017.

\bibitem{khan2021leveraging}
T.~M. Khan, S.~S. Naqvi, and E.~Meijering, ``Leveraging image complexity in
  macro-level neural network design for medical image segmentation,''
  \emph{arXiv preprint arXiv:2112.11065}, 2021.

\bibitem{khan2022t}
T.~M. Khan, A.~Robles-Kelly, and S.~S. Naqvi, ``{T-Net: A Resource-Constrained
  Tiny Convolutional Neural Network for Medical Image Segmentation},'' in
  \emph{Proceedings of the IEEE/CVF Winter Conference on Applications of
  Computer Vision}, 2022, pp. 644--653.

\bibitem{deng2009imagenet}
J.~Deng, W.~Dong, R.~Socher, L.-J. Li, K.~Li, and L.~Fei-Fei, ``Imagenet: A
  large-scale hierarchical image database,'' in \emph{2009 IEEE conference on
  computer vision and pattern recognition}.\hskip 1em plus 0.5em minus
  0.4em\relax Ieee, 2009, pp. 248--255.

\bibitem{yu2016automated}
L.~Yu, H.~Chen, Q.~Dou, J.~Qin, and P.-A. Heng, ``Automated melanoma
  recognition in dermoscopy images via very deep residual networks,''
  \emph{IEEE transactions on medical imaging}, vol.~36, no.~4, pp. 994--1004,
  2016.

\bibitem{yang2019self}
J.~Yang, X.~Wu, J.~Liang, X.~Sun, M.-M. Cheng, P.~L. Rosin, and L.~Wang,
  ``Self-paced balance learning for clinical skin disease recognition,''
  \emph{IEEE transactions on neural networks and learning systems}, vol.~31,
  no.~8, pp. 2832--2846, 2019.

\bibitem{krizhevsky2012imagenet}
A.~Krizhevsky, I.~Sutskever, and G.~E. Hinton, ``Imagenet classification with
  deep convolutional neural networks,'' \emph{Advances in neural information
  processing systems}, vol.~25, 2012.

\bibitem{radosavovic2020designing}
I.~Radosavovic, R.~P. Kosaraju, R.~Girshick, K.~He, and P.~Doll{\'a}r,
  ``Designing network design spaces,'' in \emph{Proceedings of the IEEE/CVF
  conference on computer vision and pattern recognition}, 2020, pp.
  10\,428--10\,436.

\bibitem{belkin2019reconciling}
M.~Belkin, D.~Hsu, S.~Ma, and S.~Mandal, ``Reconciling modern machine-learning
  practice and the classical bias--variance trade-off,'' \emph{Proceedings of
  the National Academy of Sciences}, vol. 116, no.~32, pp. 15\,849--15\,854,
  2019.

\bibitem{HAN20181529}
S.~S. Han, M.~S. Kim, W.~Lim, G.~H. Park, I.~Park, and S.~E. Chang,
  ``Classification of the clinical images for benign and malignant cutaneous
  tumors using a deep learning algorithm,'' \emph{Journal of Investigative
  Dermatology}, vol. 138, no.~7, pp. 1529--1538, 2018.

\bibitem{9025648}
D.~Bisla, A.~Choromanska, R.~S. Berman, J.~A. Stein, and D.~Polsky, ``Towards
  automated melanoma detection with deep learning: Data purification and
  augmentation,'' in \emph{2019 IEEE/CVF Conference on Computer Vision and
  Pattern Recognition Workshops (CVPRW)}, 2019, pp. 2720--2728.

\bibitem{wang2016attention}
Y.~Wang, M.~Huang, X.~Zhu, and L.~Zhao, ``Attention-based lstm for aspect-level
  sentiment classification,'' in \emph{Proceedings of the 2016 conference on
  empirical methods in natural language processing}, 2016, pp. 606--615.

\bibitem{chen2016neural}
H.~Chen, M.~Sun, C.~Tu, Y.~Lin, and Z.~Liu, ``Neural sentiment classification
  with user and product attention,'' in \emph{Proceedings of the 2016
  conference on empirical methods in natural language processing}, 2016, pp.
  1650--1659.

\bibitem{xia2021attention}
H.~Xia, Y.~Luo, and Y.~Liu, ``Attention neural collaboration filtering based on
  gru for recommender systems,'' \emph{Complex \& Intelligent Systems}, vol.~7,
  no.~3, pp. 1367--1379, 2021.

\bibitem{wang2017residual}
F.~Wang, M.~Jiang, C.~Qian, S.~Yang, C.~Li, H.~Zhang, X.~Wang, and X.~Tang,
  ``Residual attention network for image classification,'' in \emph{Proceedings
  of the IEEE conference on computer vision and pattern recognition}, 2017, pp.
  3156--3164.

\bibitem{hu2018squeeze}
J.~Hu, L.~Shen, and G.~Sun, ``Squeeze-and-excitation networks,'' in
  \emph{Proceedings of the IEEE conference on computer vision and pattern
  recognition}, 2018, pp. 7132--7141.

\bibitem{zhang2019attention}
J.~Zhang, Y.~Xie, Y.~Xia, and C.~Shen, ``Attention residual learning for skin
  lesion classification,'' \emph{IEEE transactions on medical imaging},
  vol.~38, no.~9, pp. 2092--2103, 2019.

\bibitem{he2022deep}
X.~He, Y.~Wang, S.~Zhao, and C.~Yao, ``Deep metric attention learning for skin
  lesion classification in dermoscopy images,'' \emph{Complex \& Intelligent
  Systems}, vol.~8, no.~2, pp. 1487--1504, 2022.

\bibitem{he2020cabnet}
A.~He, T.~Li, N.~Li, K.~Wang, and H.~Fu, ``Cabnet: category attention block for
  imbalanced diabetic retinopathy grading,'' \emph{IEEE Transactions on Medical
  Imaging}, vol.~40, no.~1, pp. 143--153, 2020.

\bibitem{tschandl2018ham10000}
P.~Tschandl, C.~Rosendahl, and H.~Kittler, ``The ham10000 dataset, a large
  collection of multi-source dermatoscopic images of common pigmented skin
  lesions,'' \emph{Scientific data}, vol.~5, no.~1, pp. 1--9, 2018.

\bibitem{combalia2019bcn20000}
M.~Combalia, N.~C. Codella, V.~Rotemberg, B.~Helba, V.~Vilaplana, O.~Reiter,
  C.~Carrera, A.~Barreiro, A.~C. Halpern, S.~Puig \emph{et~al.}, ``Bcn20000:
  Dermoscopic lesions in the wild,'' \emph{arXiv preprint arXiv:1908.02288},
  2019.

\bibitem{codella2018skin}
N.~C. Codella, D.~Gutman, M.~E. Celebi, B.~Helba, M.~A. Marchetti, S.~W. Dusza,
  A.~Kalloo, K.~Liopyris, N.~Mishra, H.~Kittler \emph{et~al.}, ``Skin lesion
  analysis toward melanoma detection: A challenge at the 2017 international
  symposium on biomedical imaging (isbi), hosted by the international skin
  imaging collaboration (isic),'' in \emph{2018 IEEE 15th international
  symposium on biomedical imaging (ISBI 2018)}.\hskip 1em plus 0.5em minus
  0.4em\relax IEEE, 2018, pp. 168--172.

\bibitem{huang2017densely}
G.~Huang, Z.~Liu, L.~Van Der~Maaten, and K.~Q. Weinberger, ``Densely connected
  convolutional networks,'' in \emph{Proceedings of the IEEE conference on
  computer vision and pattern recognition}, 2017, pp. 4700--4708.

\bibitem{Linetal2020}
T.-Y. Lin, P.~Goyal, R.~Girshick, K.~He, and P.~Dollár, ``Focal loss for dense
  object detection,'' \emph{IEEE Transactions on Pattern Analysis and Machine
  Intelligence}, vol.~42, no.~2, pp. 318--327, 2020.

\bibitem{selvaraju2017grad}
R.~R. Selvaraju, M.~Cogswell, A.~Das, R.~Vedantam, D.~Parikh, and D.~Batra,
  ``Grad-cam: Visual explanations from deep networks via gradient-based
  localization,'' in \emph{Proceedings of the IEEE international conference on
  computer vision}, 2017, pp. 618--626.

\bibitem{kassem2020skin}
M.~A. Kassem, K.~M. Hosny, and M.~M. Fouad, ``Skin lesions classification into
  eight classes for isic 2019 using deep convolutional neural network and
  transfer learning,'' \emph{IEEE Access}, vol.~8, pp. 114\,822--114\,832,
  2020.

\bibitem{grandini2020metrics}
M.~Grandini, E.~Bagli, and G.~Visani, ``Metrics for multi-class classification:
  an overview,'' \emph{arXiv preprint arXiv:2008.05756}, 2020.

\bibitem{alirezazadeh2022improving}
P.~Alirezazadeh, M.~Schirrmann, and F.~Stolzenburg, ``Improving deep
  learning-based plant disease classification with attention mechanism,''
  \emph{Gesunde Pflanzen}, pp. 1--11, 2022.

\bibitem{yu2021application}
S.~Yu, S.~Jin, J.~Peng, H.~Liu, and Y.~He, ``Application of a new deep learning
  method with cbam in clothing image classification,'' in \emph{2021 IEEE
  International Conference on Emergency Science and Information Technology
  (ICESIT)}.\hskip 1em plus 0.5em minus 0.4em\relax IEEE, 2021, pp. 364--368.

\bibitem{li2022eca}
X.~Li, H.~Xia, and L.~Lu, ``Eca-cbam: Classification of diabetic retinopathy:
  Classification of diabetic retinopathy by cross-combined attention
  mechanism,'' in \emph{2022 the 6th International Conference on Innovation in
  Artificial Intelligence (ICIAI)}, 2022, pp. 78--82.

\bibitem{woo2018cbam}
S.~Woo, J.~Park, J.-Y. Lee, and I.~S. Kweon, ``Cbam: Convolutional block
  attention module,'' in \emph{Proceedings of the European conference on
  computer vision (ECCV)}, 2018, pp. 3--19.

\bibitem{datta2021soft}
S.~K. Datta, M.~A. Shaikh, S.~N. Srihari, and M.~Gao, ``Soft attention improves
  skin cancer classification performance,'' in \emph{Interpretability of
  Machine Intelligence in Medical Image Computing, and Topological Data
  Analysis and Its Applications for Medical Data}.\hskip 1em plus 0.5em minus
  0.4em\relax Springer, 2021, pp. 13--23.

\bibitem{nozdryn2018ensembling}
A.~Nozdryn-Plotnicki, J.~Yap, and W.~Yolland, ``Ensembling convolutional neural
  networks for skin cancer classification,'' \emph{International Skin Imaging
  Collaboration (ISIC) Challenge on Skin Image Analysis for Melanoma Detection.
  MICCAI}, 2018.

\bibitem{gessert2018skin}
N.~Gessert, T.~Sentker, F.~Madesta, R.~Schmitz, H.~Kniep, I.~Baltruschat,
  R.~Werner, and A.~Schlaefer, ``Skin lesion diagnosis using ensembles,
  unscaled multi-crop evaluation and loss weighting,'' \emph{arXiv preprint
  arXiv:1808.01694}, 2018.

\bibitem{zhuang2018skin}
J.~Zhuang, W.~Li, S.~Manivannan, R.~Wang, J.~Zhang, J.~Pan, G.~Jiang, and
  Z.~Yin, ``Skin lesion analysis towards melanoma detection using deep neural
  network ensemble,'' \emph{ISIC Challenge}, vol. 2018, no.~2, pp. 1--6, 2018.

\bibitem{he2022fully}
X.~He, E.-L. Tan, H.~Bi, X.~Zhang, S.~Zhao, and B.~Lei, ``Fully transformer
  network for skin lesion analysis,'' \emph{Medical Image Analysis}, vol.~77,
  p. 102357, 2022.

\bibitem{xin2022improved}
C.~Xin, Z.~Liu, K.~Zhao, L.~Miao, Y.~Ma, X.~Zhu, Q.~Zhou, S.~Wang, L.~Li,
  F.~Yang \emph{et~al.}, ``An improved transformer network for skin cancer
  classification,'' \emph{Computers in Biology and Medicine}, p. 105939, 2022.

\bibitem{bakkouri2020computer}
I.~Bakkouri and K.~Afdel, ``Computer-aided diagnosis (cad) system based on
  multi-layer feature fusion network for skin lesion recognition in dermoscopy
  images,'' \emph{Multimedia Tools and Applications}, vol.~79, no.~29, pp.
  20\,483--20\,518, 2020.

\bibitem{zhou2019multi}
S.~Zhou, Y.~Zhuang, and R.~Meng, ``Multi-category skin lesion diagnosis using
  dermoscopy images and deep cnn ensembles,'' \emph{DysionAI, Tech. Rep}, 2019.

\bibitem{benyahia2022multi}
S.~Benyahia, B.~Meftah, and O.~L{\'e}zoray, ``Multi-features extraction based
  on deep learning for skin lesion classification,'' \emph{Tissue and Cell},
  vol.~74, p. 101701, 2022.

\bibitem{villa2022optimized}
J.~P. Villa-Pulgarin, A.~A. Ruales-Torres, D.~Arias-Garzon, M.~A. Bravo-Ortiz,
  H.~B. Arteaga-Arteaga, A.~Mora-Rubio, J.~A. Alzate-Grisales, E.~Mercado-Ruiz,
  M.~Hassaballah, S.~Orozco-Arias \emph{et~al.}, ``Optimized convolutional
  neural network models for skin lesion classification,'' \emph{Comput. Mater.
  Contin}, vol.~70, no.~2, pp. 2131--2148, 2022.

\end{thebibliography}
\end{document}